\pdfoutput=1
\documentclass[final,5p,times,twocolumn,authoryear]{elsarticle}

\usepackage{amssymb}
\usepackage{listings}
\usepackage{booktabs} % for professional tables
\usepackage{adjustbox}
\usepackage{url}
\usepackage{hyperref}
\usepackage{multirow}
\usepackage{amsmath}

\newcommand\tabVSpace{0.2cm}

\journal{}

\begin{document}

\begin{frontmatter}



\title{Deep Learning Computer Vision Algorithms for\\Real-time UAVs On-board Camera Image Processing}


\author{Alessandro Palmas - Nurjana Technologies srl - \texttt{alessandro.palmas@nurjanatech.com}\\Pietro Andronico - Nurjana Technologies srl - \texttt{pietro.andronico@nurjanatech.com}}

\begin{abstract}
This paper describes how advanced deep learning based computer vision algorithms are applied to enable real-time on-board sensor processing for small UAVs. Four use cases are considered: target detection, classification and localization, road segmentation for autonomous navigation in GNSS-denied zones, human body segmentation, and human action recognition. All algorithms have been developed using state-of-the-art image processing methods based on deep neural networks. Acquisition campaigns have been carried out to collect custom datasets reflecting typical operational scenarios, where the peculiar point of view of a multi-rotor UAV is replicated. Algorithms architectures and trained models performances are reported, showing high levels of both accuracy and inference speed. Output examples and on-field videos are presented, demonstrating models operation when deployed on a GPU-powered commercial embedded device (NVIDIA Jetson Xavier) mounted on board of a custom quad-rotor, paving the way to enabling high level autonomy.

\end{abstract}




\end{frontmatter}


\section{Introduction}
\label{introduction}

Recent trends in the domain of unmanned aerial vehicles (UAVs) see the extensive exploration of drone swarms solutions, especially leveraging the increasing availability of small systems, with both rotating and fixed wings architectures, at low costs. These new advancements allow the design and execution of new types of missions, being able to exploit a different configuration of tactical assets, and the increased stream of data made available by the high number of sensors that can be deployed. 

In this new setting, additional bottlenecks arise. In fact, every sensor on-board the flying assets requires, in the majority of cases, a dedicated operator to continuously monitor it. This poses an important limitation to the scalability of the system, depending on the available qualified manpower.

In the last decade, another field has seen a relevant growth in terms of technological capabilities, opening major opportunities. The computer vision domain underwent a revolution thanks to the breakthroughs achieved in the deep learning (DL) field. Performances of image processing algorithms overpassed every established state-of-the-art reference, unlocking applications that were not possible before.

This paper describes how advanced deep learning based computer vision algorithms are applied to enable real-time on-board sensor processing for small UAVs. Four use cases are considered: target detection, classification and localization, road segmentation for autonomous navigation in GNSS-denied zones, human body segmentation, and human action recognition. 

The first step has been a thorough literature review of deep learning methods for image processing, in order to identify the best performing solution for each of the four domains. The models have been selected focusing on both accuracy and speed, given the need of providing real-time inference.

All algorithms have been developed using state-of-the-art image processing methods based on deep neural networks. Acquisition campaigns have been carried out to collect custom datasets reflecting typical operational scenarios, where the peculiar point of view of a multi-rotor UAV is replicated. Acquired images have been manually annotated to be used for training and evaluation. 

Selected models can be applied on RGB images as demonstrated, as well as on gray scale, infra-red, multi-spectral and hyper-spectral data. In addition, the system can be easily customized to be trained on custom targets. 

They have been trained applying all relevant best practices for deep learning training. Obtained performances are reported, demonstrating a high level of both accuracy and inference speed. Finally, output images and videos references are shared, showing models operation when deployed on a GPU-powered commercial embedded device (NVIDIA Jetson Xavier) mounted on board of a custom quad-rotor, paving the way to enabling high level autonomy.

The paper is organized as follows: Section \ref{background} provides a detailed literature review that presents state-of-the-art models and the ones that have been selected, Section \ref{training-performance-deployment} discusses model training and deployment details, Section \ref{results} presents the results that have been achieved and in Section \ref{conclusions} is reported a discussion with conclusions.

\section{Background and Literature Review}
\label{background}

In the last ten years the computer vision research has progressed at unprecedented pace, mainly thanks to the rise of the deep learning technology and, in particular, convolutional neural networks (CNN). Their performance in the context of image processing has overpassed every previous state-of-the-art record, making them the undiscussed best technical solution for these tasks \citep{Wu20}.

\subsection{Object Detection, Classification and Localization}

When focusing on DL algorithms for object detection, classification and localization, two main families of methods can be identified \citep{Sultana20} \citep{Jiao19}: two-stages approaches, as for example the R-CNN series \citep{He17} \citep{Girshick14} \citep{Li17} \citep{Ren15}, and one-stage approaches, as for example SSD \citep{Liu16} and YOLO \citep{Redmon16}. These two categories can be defined respectively as region proposal based and regression/classification based \citep{Zhao19} \citep{Liu21}. The former can be considered a traditional approach, in which regions of interest (ROIs) are generated in a first passage and then, in a second step, they are processed/classified. In the latter, instead, the localization and classification operations are performed in a unique passage. One-stage methods are in general faster, while two-stages ones are generally more accurate \citep{Soviany29}. In what follows the most important examples of the two groups are briefly described.

\subsubsection{Two-stages Approaches}

The main idea at the core of the two-stages family is the generation of a great number of ROIs in the image, so that every object in the scene is present in one or more of them. In the second stage, each of these ROIs is finally classified. The most important algorithms among this category are: R-CNN \citep{Girshick14} (2014), one of the first to beat classical methods such as HOG descriptor-based, but still too high in computational cost \citep{Sharma20}. SPP-net \citep{He14} (2014), that couples the CNN with a Spatial Pyramid Pooling (SPP) level for features extraction, extending applicability to images of different sizes and achieving a 20x speedup, while not being able to undergo an end-to-end training as SPP does not support back-propagation. Fast R-CNN \citep{Girshick15} (2015) substitutes the SPP layer with a ROI pooling layer, which supports back-propagation, thus making end-to-end training possible. Faster R-CNN \citep{Ren15} (2015) extends the last one leveraging a Region Proposal Network (RPN) to generate ROIs of different sizes and proportions, improving efficiency and obtaining a 10x speedup with respect to the Fast-R-CNN \citep{Jiao19} \citep{Simonyan15}.

\subsubsection{One-stage Approaches}

These methods skip the ROIs generation step typical of two-stages approaches, considering all subregions of the image as candidates objects. This sensibly reduces processing time and makes these methods better suited for real-time applications. The most important algorithms among this category are: You Only Look Once (YOLO) \citep{Redmon16} (2015), the first to work in real-time while maintaining a good accuracy, it also features a simplified version (Tiny-YOLO) for inference speed-up. It generates potential ROIs by dividing the image in a NxN grid. Its performances degrade on objects that are small with respect to the image size and on those with relevant occlusion. YOLO underwent a series of improvements, leading to YOLO9000 \citep{Redmon17} (2016), YOLOv3 \citep{Redmon18} (2018) and YOLOv4 \citep{Bochkovskiy20} (2020). Modifications implemented dealt with the neural network architecture, its pre-training, a new automated anchors system, the integration of the batch-normalization technique and multi-scale training. They allowed YOLO to become a top performing model in terms of both accuracy and inference speed. Single Shot Detector (SSD) \citep{Liu16} (2016) directly predicts the object class and performs bounding box regression operating at different levels, achieving competitive results, in particular for the SSD512 model, using VGG-16 as convolutional backbone network. RetinaNet \citep{Lin17b} (2017) features a loss function aiming at reducing the "foreground-background class imbalance" \citep{Chen20a} allowing it to obtain an accuracy and speed higher than those of the two-stages method already described.

\subsection{Semantic Segmentation}

These models have the goal of generating a pixel-map for a given image, where a class label is assigned to each pixel in the input. The most important families of this type of models are presented in this section.

Fully Convolutional Networks (FCN) \citep{Long15} have been the first methods used for deep learning-based image segmentation, using only convolutional layers. A specific variant, FCN-8s, has been among the first to obtain state-of-the-art-results in segmentation tasks, while not being suited for real-time application due to high inference times, and not efficient in leveraging context information.  ParseNet \citep{Liu15b} extends FCNs concatenating latent features from the CNN with a vector of global information, obtaining sensible improvement in model performance.

To address the problem of information loss, deep learning-based models have been coupled with probabilistic ones, as the Conditional Random Field (CRFs) and Markov Random Field (MRF). DeepLab-CRF (DeepLabv1) \citep{Chen14} puts together a CNN and a CRF model that processes the output of the former, providing a more accurate segmentation map. Deep Parsing Network (DPN) \citep{Liu15a}, instead, leverages MRF in a similar fashion, obtaining a relevant improvement in terms of both accuracy and inference time.

The encoder-decoder family uses convolutional layers to generate a compressed input representation in the latent space, and deconvolutional ones to reconstruct the image. One of the main problems of these models is setting the right level of abstraction, limiting information loss and the impact on achievable accuracy. DeconvNet \citep{Noh15} (2015) and SegNet \citep{Badrinarayanan17} (2017) both use VGG-16, without the last FC layers, as encoder, but the latter uses additional connections between encoding-decoding pooling-up-sampling layers at different depth, resulting in better performance in both accuracy and inference speed. Efficient Symmetric Network (ESNet) \citep{Wang19} is an encoder-decoder network optimized for real-time applications but maintaining a very good accuracy. U-Net \citep{Ronneberger15} features some additional direct connections between encoding and decoding layers and, while being conceived in the context of biomedical applications, it has also been applied in other domains, such as road-segmentation applications \citep{Zhang18b}.

Another family of segmentation models is based on multi-scale pyramids and parallel paths. Feature Pyramid Network (FPN) \citep{Lin17a} (2016) creates connections between high and low-resolution layers with a minimal computational cost. Pyramid Scene Parsing Network (PSPN) \citep{Zhao17} takes into account the global context more accurately, applying pooling operations on different levels of the layers hierarchy. High-Resolution Network (HRNet) \citep{Sun19} and Object-Contextual Representation (OCR) \citep{Yuan20} (2019) extract information on different layers with connections favoring exchange between different hierarchy levels. Deep Dual-Resolution Network (DDRNet) \citep{Hong21} (2021), based on HRNet, obtained a relevant improvement in terms of inference speed. The network has two different branches: one extracts a feature map at high resolution, while the other extracts context information using down-sampling. FasterSeg \citep{Chen20b}, inspired by Auto-DeepLab \citep{Liu19}, makes use of parallel branches too, showing a good compromise in terms of accuracy and inference speed. 

A different family of models for semantic segmentation is based on dilated convolution, a modified version of the standard convolution, allowing to reduce the number of model parameters, and thus computational cost, while obtaining similar results, fostering real-time applications. DeepLabv2 \citep{Chen18a} uses the dilated convolution, it introduces a module called atrous spatial pyramid pooling (ASPP) to capture both the global context and the low level information, and it couples the CNN with a complementary model as seen in the previous DeepLab model. Updated versions have been proposed after it, DeepLabv3 \citep{Chen18b} and DeepLabv3+ \citep{Liu19}, where the latter builds upon the former, using it as encoder in the encoder-decoder fashion. Efficient Network (ENet) \citep{Paszke16}, is specifically designed for real-time applications and leverages the lower computational requirements of the dilated convolution in the encoder-decoder context, showing a specific feature of having an asymmetrical encoding-decoding branches, with a deeper encoder and a more shallow decoder. Gated Shape CNN (GSCNN) \citep{Takikawa19} has a "regular stream", where a CNN generates feature maps, and a "shape stream", where a gated convolutional layer (GCL) uses CNN output to more accurately deal with objects contours. The output is then fused in an ASPP module where contextual information is preserved on different scales.

\subsection{Human Action Recognition}

The aim of these algorithms is to classify a sequence of images in a given set of categories. Typical examples of labels can be "walking", "running", "standing idle" or "throwing something". This section presents the most important models available in the literature based on visual input data. They are typically divided in two categories: methods based on two or more 2D fluxes and methods based on recurrent neural networks (RNN).

\subsubsection{Methods Based on Two or More 2D Fluxes}

These methods make use of at least two 2D CNNs working in parallel. They work as independent features extractors operating on the input images sequence, and their classification outputs are then combined to produce the final result. Two-Stream CNN+SVM \citep{Simonyan14} (2014) and Multi-Resolution CNN \citep{Karpathy14} (2014) both feature two branches operating, respectively, on a single frame and the optical-flow obtained from a sequence of multiple frames, and low and high resolution frames. The first fuses the classification output of each branch, while the second directly concatenates the convolutional layers output. Trajectory-pooled Deep-convolutional Descriptors (TDD) \citep{Wang15} (2015) uses two parallel fluxes based on deep learning and hand-crafted operations respectively, then using "Fisher Vector" representation \citep{Sanchez13} as input to a classifying SVM. Temporal Segment Network (TSN) \citep{Wang16} (2016) has been designed to address complex actions covering a large time span, it divides the input frame sequence in three parts and forward them to three networks, fusing their output for classification. Temporal Linear Encoding (TLE) \citep{Diba17} aims at recognizing long-duration actions too and features a specific layer for this purpose. ActionVLAD \citep{Girdhar17} (2017) processes in parallel RGB and flow streams and then applies a particular pooling layer embedding a clustering operation where classification is carried out through proximity measures. Two-stream ConvNet \citep{Feichtenhofer17} (2017) introduced spatial-temporal information fusion in the convolutional layers instead of limiting it to outputs, resulting in improved performances and in a lower number of parameters. CNN + Deep AutoEncoder + Support-Vector-Machine (CNN + DAE + SVM) \citep{Ullah19} (2019) processes videos in quasi-real-time while maintaining the same accuracy of the previous ones. The outputs of the two parallel features extractors is processed by FC layers, an auto-encoder, and finally a SVM generating the classification. MSM-ResNet \citep{Ming21} (2021) uses three parallel CNN-based processing pipelines, increasing computational costs, operating on the single frame, the optical flow and motion saliency stream, and all three are used to generate the final prediction.

\subsubsection{Methods Based on Recurrent Neural Networks}

These models try to solve the limits encountered when dealing with actions of long durations, that extend beyond the number of frames the model processes at the same time. They leverage Recurrent Neural Networks (RNN), modules having neurons linked in loop. This generates a "memory" effect, allowing to handle tasks requiring to take decisions based on information received far back in time, which makes them suitable for dynamic applications like action recognition. This section presents these models, having a Long-Short-Term-Memory (LSTM) block (a specific RNN module) which processes features extracted by two or more parallel CNNs. The additional block results in additional computational costs with respects to the previous ones. Long-term Recurrent Convolutional Networks (LRCNs) \citep{Donahue17} (2015) uses multiple CNN-based features extractors followed by LSTM blocks and an averaging module that performs the final classification. Lattice Long-Short-Term Memory (L2STM) \citep{Sun17} (2017) features two CNN-based features extractors and a Lattice-LSTM module, which extends the classic LSTM, allowing for cross flux information exchange. ShuttleNet \citep{Shi17} (2017) does not use LSTM blocks as recurrent units, maintaining the computational cost on par with the other models. Attention Mechanism-Long Short Time Memory (AM-LSTM) \citep{Ge19} (2019) can be decomposed in four main modules: a CNN-based feature extractor, a module based on the attention mechanism with the purpose of automatically select the most important features extracted, a RNN module based on convolutional LSTM (ConvLSTM) that generates predictions based on the frame, and a final prediction block providing the classification output. Densely-connected Bi-directional Long-Short Time Memory (DB-LSTM) \citep{He21} (2021) creates batches sampling from the input frames sequence and optical flux, and sends them to the sample representation learned (SRL) block, composed by two parallel processing pipelines, one for spatial and one for temporal information, leveraging CNNs for features extraction. The output is provided to the bidirectional LSTMs that, given this peculiar feature, allow to capture information for longer time-frames with respect to conventional LSTMs. The last element is a fusion layer that generates the classification.

\subsection{Model Selection}

Following the literature review, a model for each of the four use cases has been identified: YOLOv3 \citep{Redmon18} for object detection, classification and localization, DDRNet \citep{Hong21} for the two segmentation scenarios (road segmentation and human body segmentation), and Two-stream ConvNet \citep{Feichtenhofer17} for human action recognition.

All the algorithms have been customized to easily deal with different type of inputs, supporting RGB and gray-scale images in the visible spectrum, gray-scale images in the infra-red spectrum, and images acquired by multi-spectral cameras or (typically satellite based) hyper-spectral sensors. These modifications have also taken into account the specific deployment hardware used for on-board application, and the need for quasi-real-time inference.

\section{Model Training, Deployment and Performance}
\label{training-performance-deployment}

With the aim of carrying out on-field demonstrations, we focused on specific applications in the context of the selected use cases. Object detection, classification and localization is demonstrated on a single class (person) application, using RGB images acquired from a drone flying 10 meters above the ground with the camera oriented downwards (-90 $\deg$). This scenario has been chosen being it representative of situations where, for example, there's the need to detect a target and autonomously track it in the contexts of search \& rescue missions or restricted area monitoring.

The model for semantic segmentation has been applied in two different contexts. The first aims at distinguishing between road and off-road pixels in RGB images acquired from a drone flying 10 meters above the ground with the camera oriented downwards (-90 $\deg$). This functionality can be used to implement autonomous navigation, particularly useful in GNSS-denied zones, to build for example a system able to autonomously follow a road for patrolling operations.

The second application of semantic segmentation aimed at segmenting the human silhouette in 19 different body parts (like head, hair, arms, legs, torso, feet) in RGB images acquired from a drone flying at 5 meters above the ground with the camera oriented at -40 $\deg$. 

The model for human action recognition has been trained to classify six different actions in RGB frames sequences acquired from a drone flying 10 meters above the ground with the camera oriented downwards (-90 $\deg$). Actions performed were: standing idle, walk, run, crouch, aim and throw. This feature can be used, for example, in crowded areas monitoring applications to automatically identify threatening behavior.

\subsection{Dataset}
\label{dataset}

\begin{figure}[h]
\vskip 0.2in
\begin{center}
\centerline{\includegraphics[width=\columnwidth]{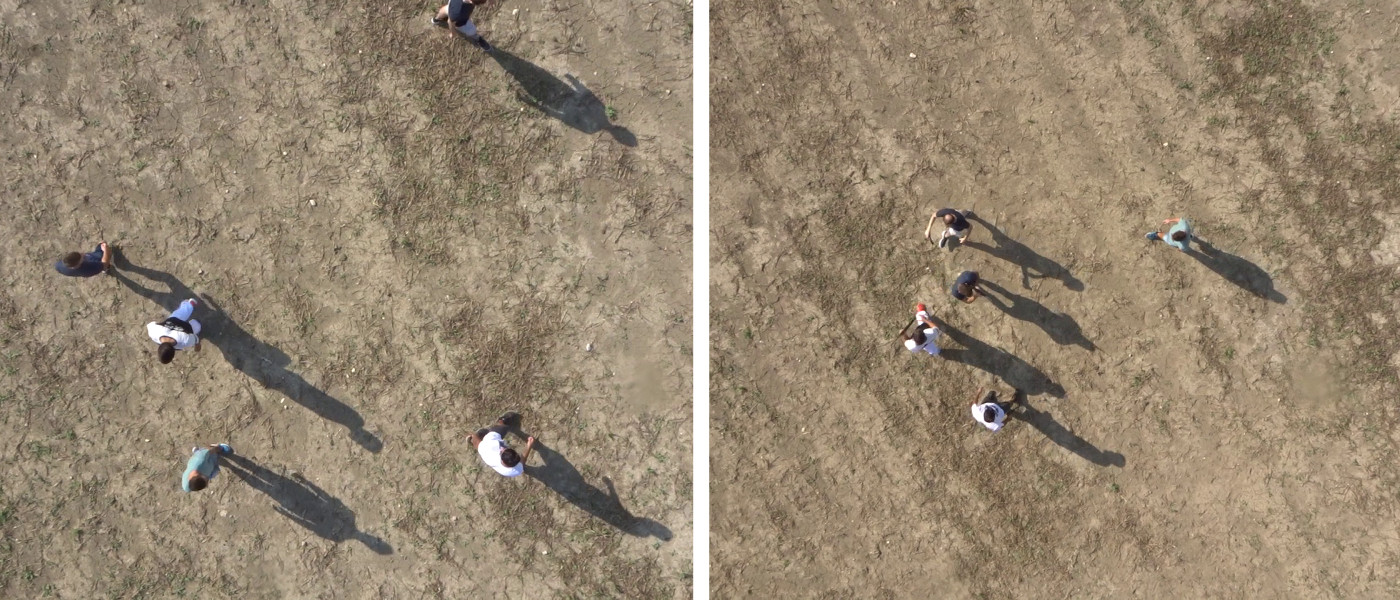}}
\caption{Samples of the dataset used for the object detection, classification and localization scenario}
\label{datasetOd}
\end{center}
\vskip -0.2in
\end{figure}

\begin{figure}[h]
\vskip 0.2in
\begin{center}
\centerline{\includegraphics[width=\columnwidth]{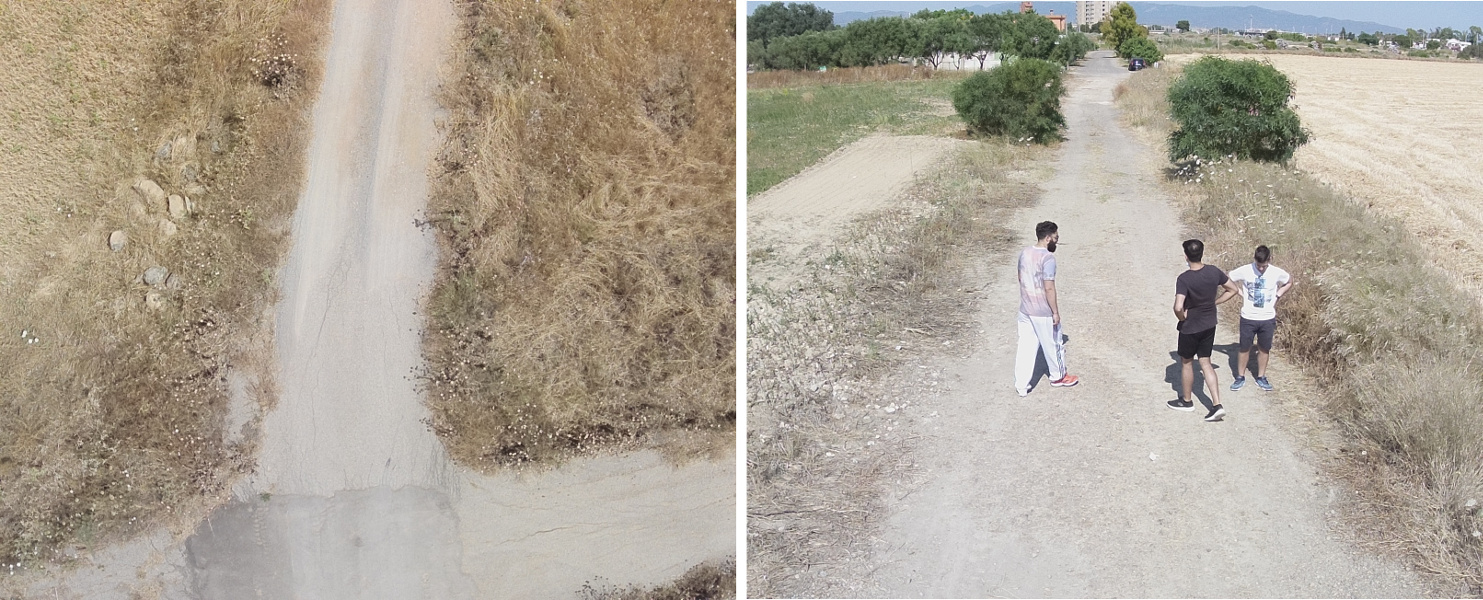}}
\caption{Samples of the dataset used for road segmentation (left) and human body segmentation (right) scenarios}
\label{datasetSeg}
\end{center}
\vskip -0.2in
\end{figure}

\begin{figure}[h]
\vskip 0.2in
\begin{center}
\centerline{\includegraphics[width=\columnwidth]{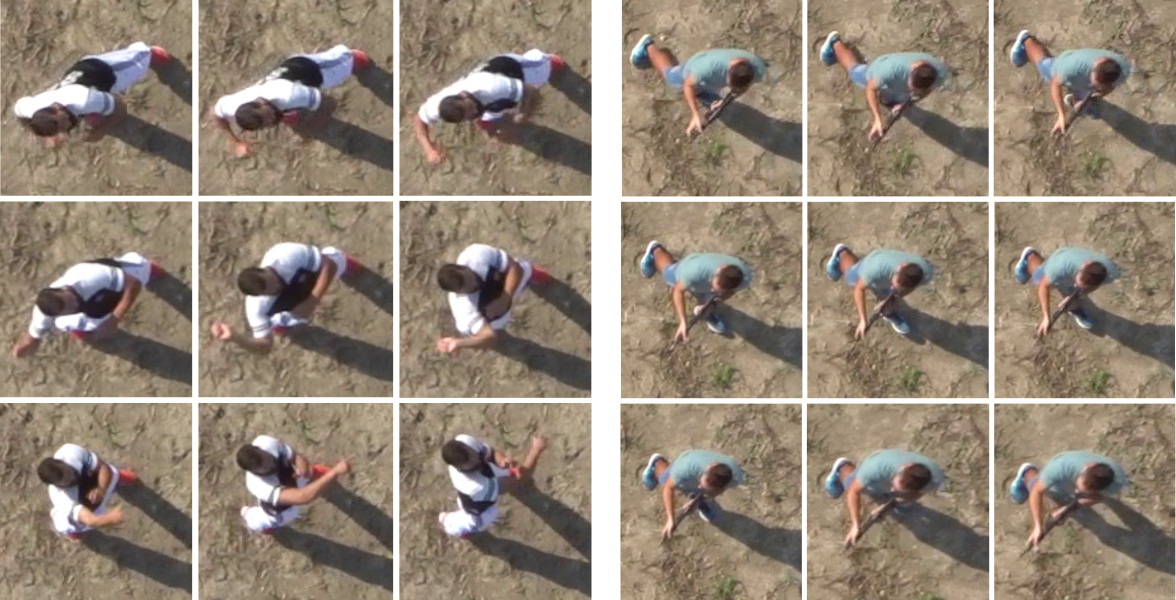}}
\caption{Samples of the dataset used for human action recognition scenario (throwing action on the left, aiming action on the right).}
\label{datasetHar}
\end{center}
\vskip -0.2in
\end{figure}

The four use cases considered in this study, and the goal of demonstrating their application in real world tests after having deployed them on-board of a multi-rotor drone, posed a significant challenge. In fact, while a relevant number of labeled image datasets are openly available for all the different applications here considered, only a few of them feature the very peculiar airborne point of view needed in our case. For this reason, we carried out a number of on-field campaigns with the aim of building our own custom dataset. \textbf{Figures \ref{datasetOd}, \ref{datasetSeg} and \ref{datasetHar}} show some samples used for object detection, segmentation and action recognition respectively.

\begin{figure}[ht]
\vskip 0.2in
\begin{center}
\centerline{\includegraphics[width=\columnwidth]{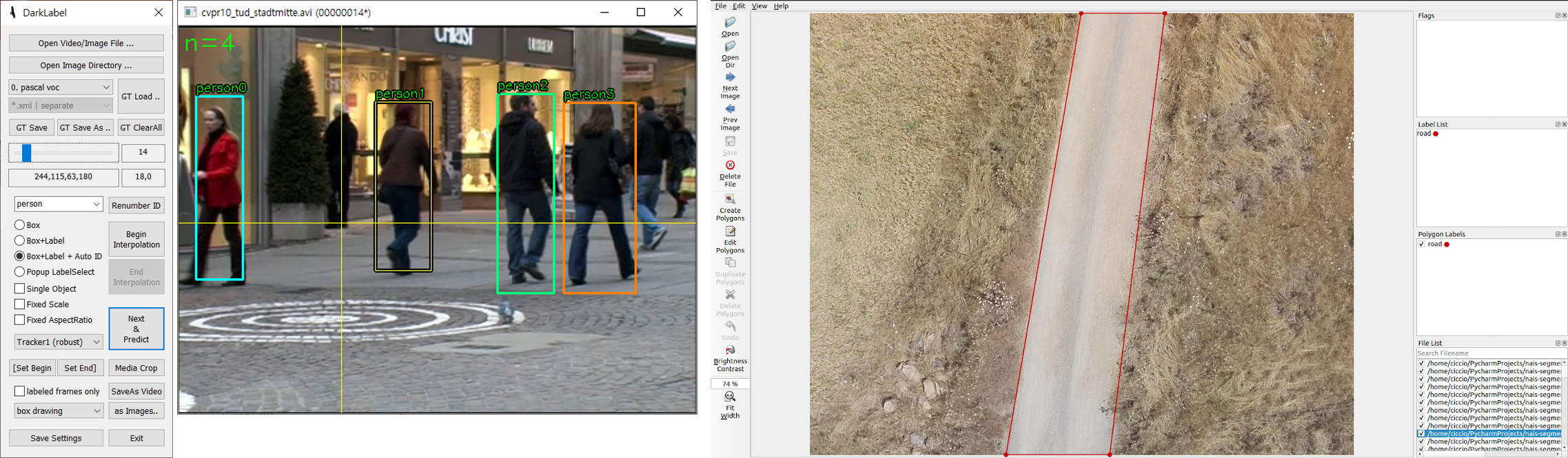}}
\caption{Annotation tool used for object detection and action recognition (left) and semantic segmentation (right)}
\label{annotation1}
\end{center}
\vskip -0.2in
\end{figure}

After the collection campaigns, we manually labeled the data using the tools presented in \textbf{Figure \ref{annotation1}}.

\begin{figure}[ht]
\vskip 0.2in
\begin{center}
\centerline{\includegraphics[width=\columnwidth]{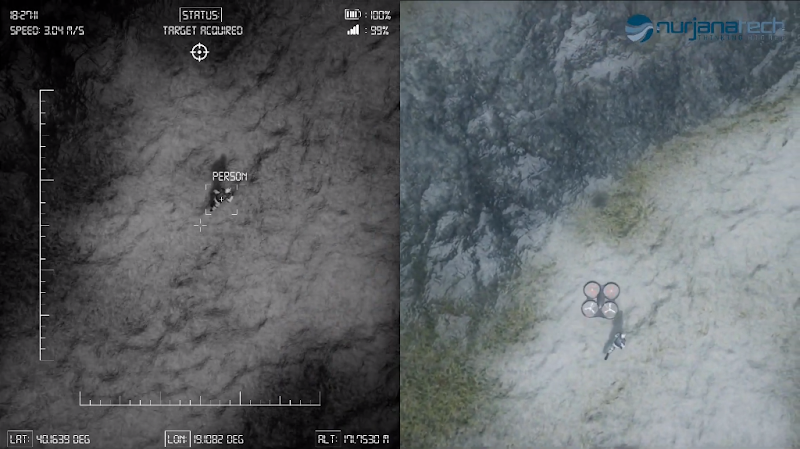}}
\caption{Application of the object detection algorithm in a virtual scenario generated with Unreal Engine}
\label{synthetic}
\end{center}
\vskip -0.2in
\end{figure}

As an additional research direction we wanted to explore, we also leveraged Unreal Engine to create synthetic datasets having similar characteristics to those used for the four considered use cases, and to virtually test the algorithms and their integration with the drone navigation dynamics. \textbf{Figure \ref{synthetic}} shows the application of object detection algorithm in a virtual scenario. Making a proper use of these engines allows to by-pass the labeling step, leveraging the knowledge of the virtual environment composition underlying the rendered scene, completely automating dataset generation. In addition, it provides a great flexibility, as the user can vary:
\begin{itemize}
\item Environmental conditions: weather (sun, rain, fog, etc.), time-of-day, environment (urban, desert, woods, rural, etc.)
\item Target types: category of assets for object detection, actions executed, assets appearance (clothes, wearables, equipment)
\item Test conditions to investigate algorithms generalization and robustness
\end{itemize}

just to list the most important ones.

\subsection{Training Techniques}
\label{training-techniques}

After having collected the dataset for all the considered use cases, the different training strategies described below have been applied. 

\textbf{Training-Validation-Test Split.} For each of the four algorithms, the dataset has been split in training, validation and test sets, making sure label balance was properly preserved. To optimize data usage, a 5-fold partitioning has been adopted, averaging scores across folds to obtain model performance estimations. 

\begin{figure}[ht]
\vskip 0.2in
\begin{center}
\centerline{\includegraphics[width=\columnwidth]{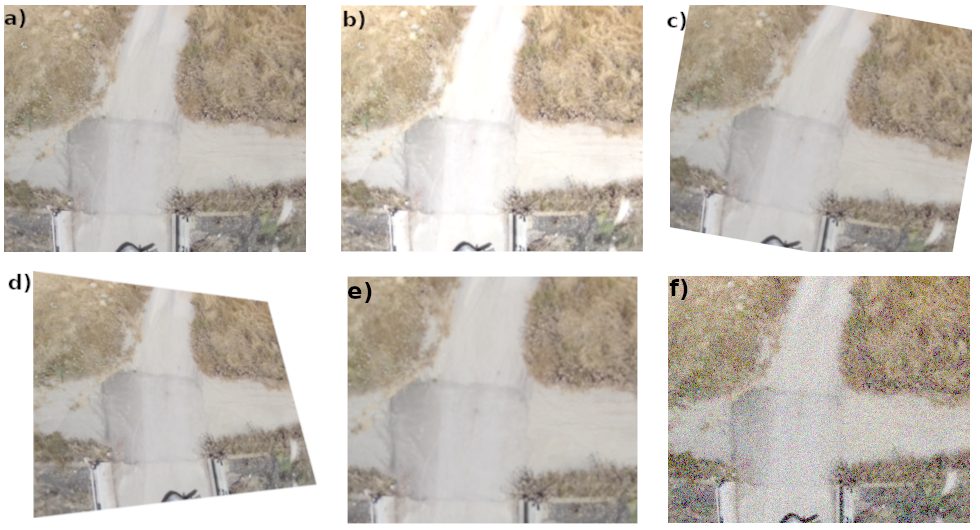}}
\caption{Data augmentation operations: a) original image b) brightness shift c) rotation d) shear affine transformation e) cropping f) noise superimposition}
\label{dataAug}
\end{center}
\vskip -0.2in
\end{figure}

\textbf{Data Augmentation.} With the aim of enhancing generalization power of the algorithms, different data augmentation techniques have been applied on the collected images. \textbf{Figure \ref{dataAug}} presents them clearly, starting from the top left and going row-wise one finds: the original image, application of a brightness shift, rotation of the image, application of a shear affine transformation, random cropping and noise superimposition.

These transformations are applied at training time, randomizing the parameters they depend on. 

\textbf{Pre-training, Layer Freezing and Fine Tuning.} All selected models have been pre-trained using openly available datasets. As a next step, a first round of training made use of the synthetic generated data, to drive learning towards the real world scenarios. 

\begin{figure}[ht]
\vskip 0.2in
\begin{center}
\centerline{\includegraphics[width=\columnwidth]{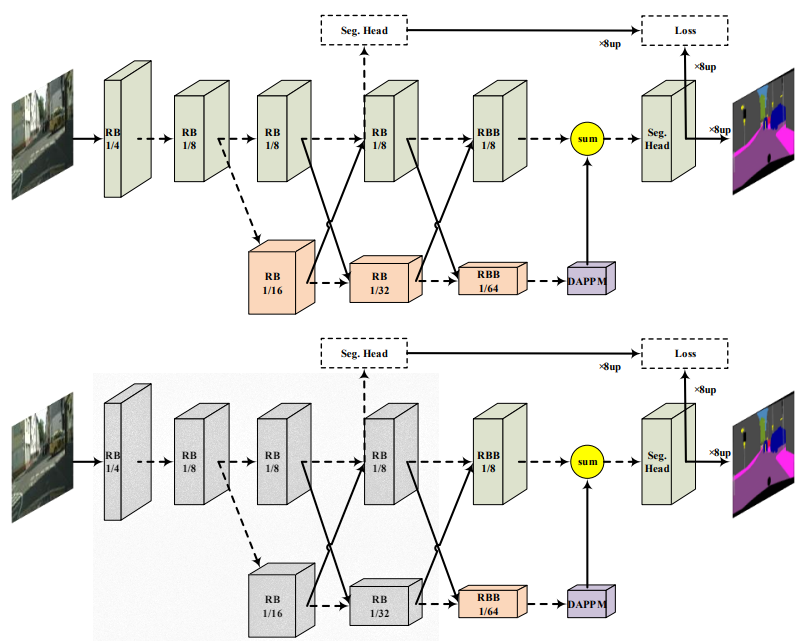}}
\caption{Layer freezing for transfer learning and fine-tuning}
\label{freezing}
\end{center}
\vskip -0.2in
\end{figure}

On the resulting models, the first layers have been frozen, \textbf{Figure \ref{freezing}} intuitively shows this on the DDRNet model where frozen layers are grayed out, and the deep network has been fine-tuned on the dataset collected in the acquisition campaigns.

Finally, early stopping based on the validation performance measure has been adopted to avoid over-fitting.

\subsection{Deployment}

All the algorithms have been implemented using state of the art deep learning frameworks (TensorFlow / Pytorch). To speed-up hyper-parameters tuning and experimentation, training has been carried out on a desktop computer powered by a NVIDIA RTX 3090 GPU. 
Deployment constraints have guided the development from the start, thus models were designed to be easily portable to embedded, GPU-powered devices. Trained algorithms have been deployed on the NVIDIA Jetson Xavier board, which is particularly well suited to be equipped on-board, due to its low weight and power consumption, as well as very flexible hardware and software interfaces. 

To maximize performances, reduced precision arithmetics have been adopted leveraging NVIDIA TensorRT tool. 

\begin{figure}[ht]
\vskip 0.2in
\begin{center}
\centerline{\includegraphics[width=\columnwidth]{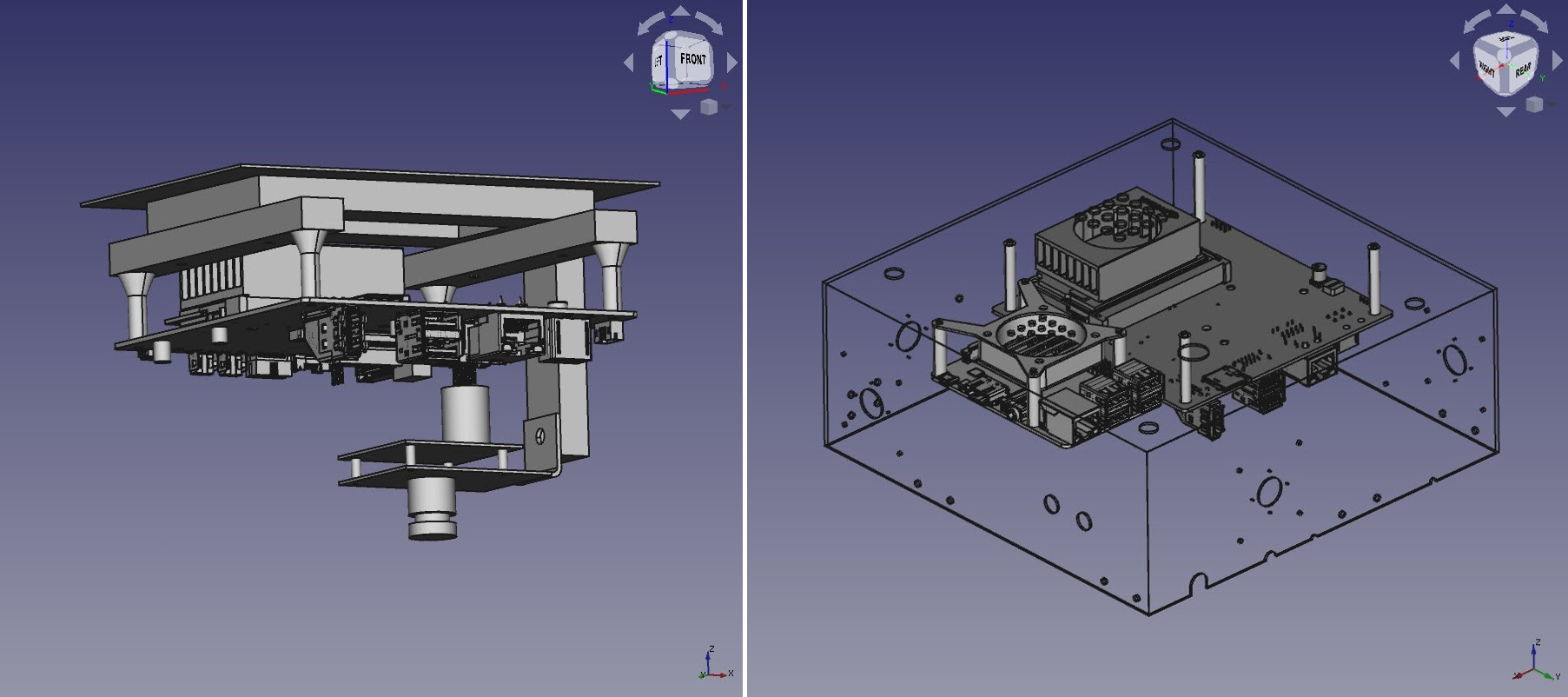}}
\caption{Custom design of NVIDIA Jetson's support for deployment on-board of a multi-rotor drone}
\label{jetsonSupport}
\end{center}
\vskip -0.2in
\end{figure}

\begin{figure}[ht]
\vskip 0.2in
\begin{center}
\centerline{\includegraphics[width=\columnwidth]{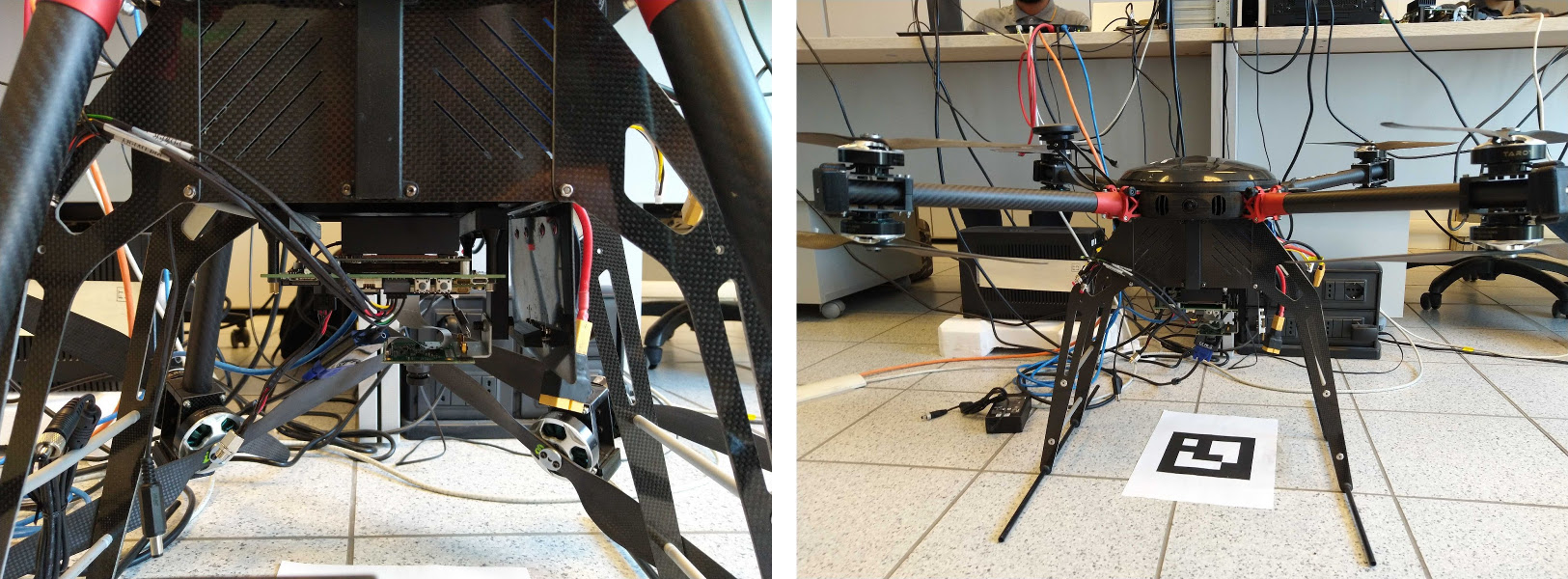}}
\caption{NVIDIA Jetson deployed on-board of a multi-rotor drone}
\label{jetsonSupport2}
\end{center}
\vskip -0.2in
\end{figure}

\textbf{Figure \ref{jetsonSupport}} shows the custom design of a mechanical interface to equip the NVIDIA hardware on-board of a multi-rotor drone as an additional payload. In \textbf{Figure \ref{jetsonSupport2}} is shown the final prototype assembled and ready to be flown for tests using a quad-x configuration opto-copter. 

\section{Results}
\label{results}

The following table summarizes the performances of implemented algorithms. For each algorithm it shows two metrics, one for inference speed and one for accuracy. The former is measured for every algorithm in terms of frames per second, while the latter depends on the specific application and presents the most meaningful metric typically used with it. 

\begin{table*}[ht]
\caption{Algorithms Performances}
\vskip 0.05in
\begin{center}
\begin{small}
\begin{sc}
\begin{adjustbox}{max width=\textwidth}
\begin{tabular}{cccc}
\toprule
Algorithm & Performance Measure & Result \\
\midrule
\multirow{2}{*}{\begin{tabular}{@{}c@{}}Automatic target detection,  \\
  classification and localization\end{tabular}} & Frames Per Second (FPS) & 15-20 FPS \\[\tabVSpace]
 & Mean Average Precision (mAP) & 70.5 \% \\
 \midrule
\multirow{2}{*}{Road Segmentation} & Frames Per Second (FPS) & 10 FPS\\[\tabVSpace]
 & Intersection over Union (IoU) & 93.86 \%\\
 \midrule
\multirow{2}{*}{Human Body Segmentation} & Frames Per Second (FPS) & 25 FPS \\[\tabVSpace]
 & Intersection over Union (IoU) & 47.84 \%\\
 \midrule
\multirow{2}{*}{Human Action Recognition} & Frames Per Second (FPS) & 10 FPS\\[\tabVSpace]
 & Mean Average Precision (mAP) & 50.8 \%\\
\bottomrule
\end{tabular}
\end{adjustbox}
\end{sc}
\end{small}
\end{center}
\vskip -0.1in
\label{algorithms-performances-table}
\end{table*}

\begin{figure}[h]
\vskip 0.2in
\begin{center}
\centerline{\includegraphics[width=\columnwidth]{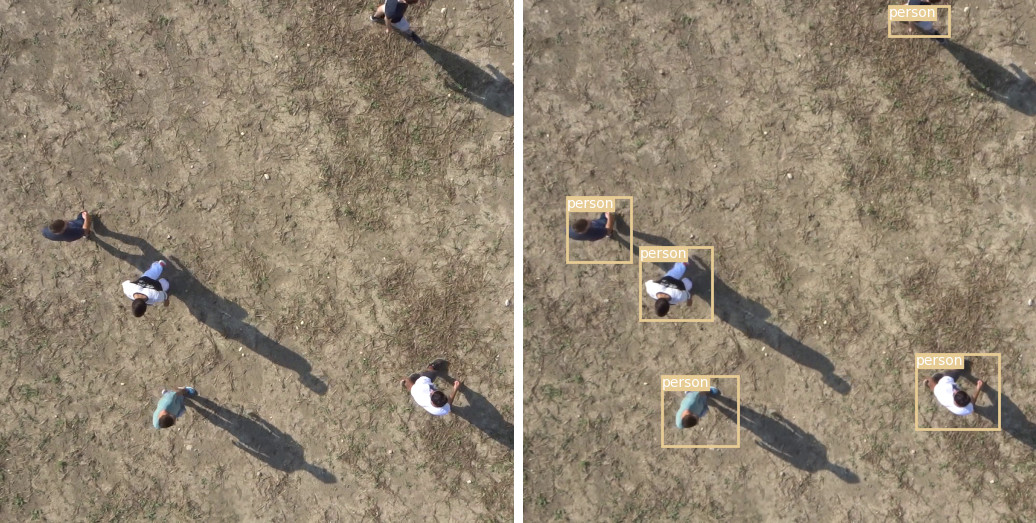}}
\caption{Example of the object detection, classification and localization model output}
\label{object-detection}
\end{center}
\vskip -0.2in
\end{figure}

\begin{figure}[ht]
\vskip 0.2in
\begin{center}
\centerline{\includegraphics[width=\columnwidth]{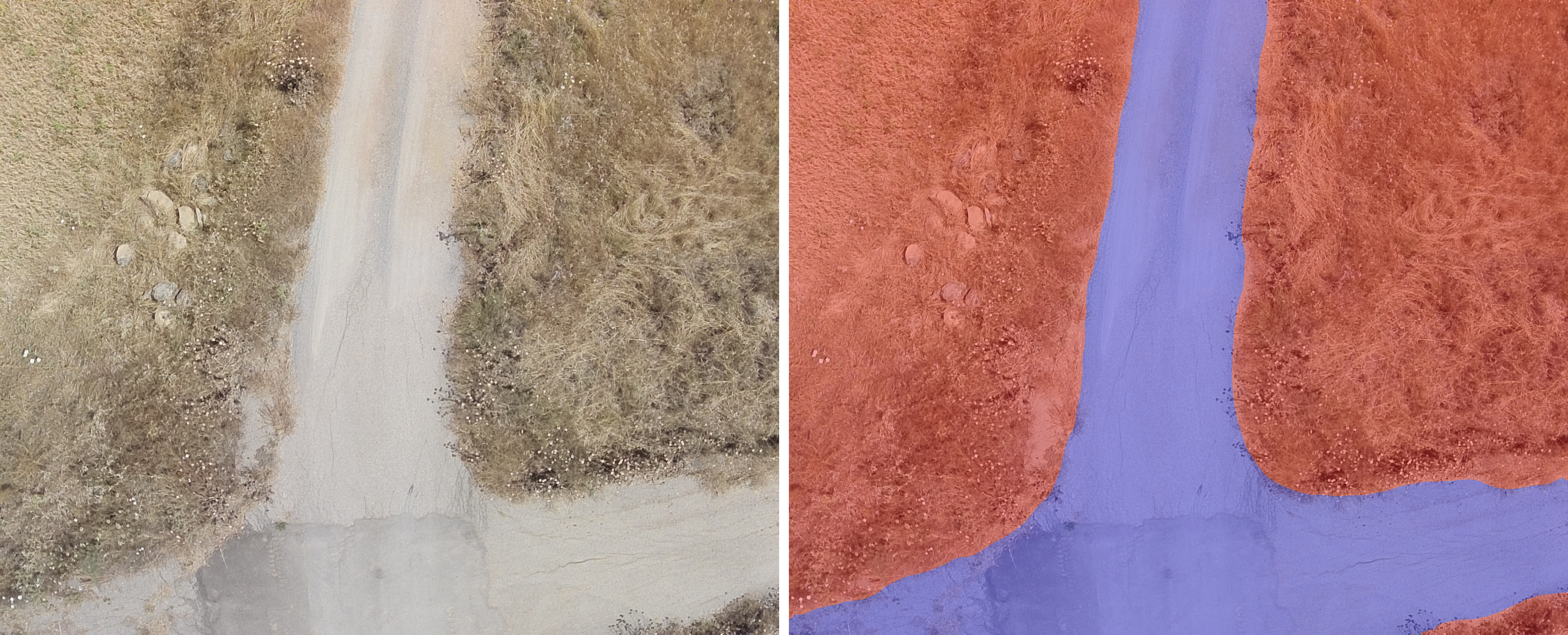}}
\caption{Example of road segmentation model output}
\label{road-seg}
\end{center}
\vskip -0.2in
\end{figure}

\begin{figure}[ht]
\vskip 0.2in
\begin{center}
\centerline{\includegraphics[width=\columnwidth]{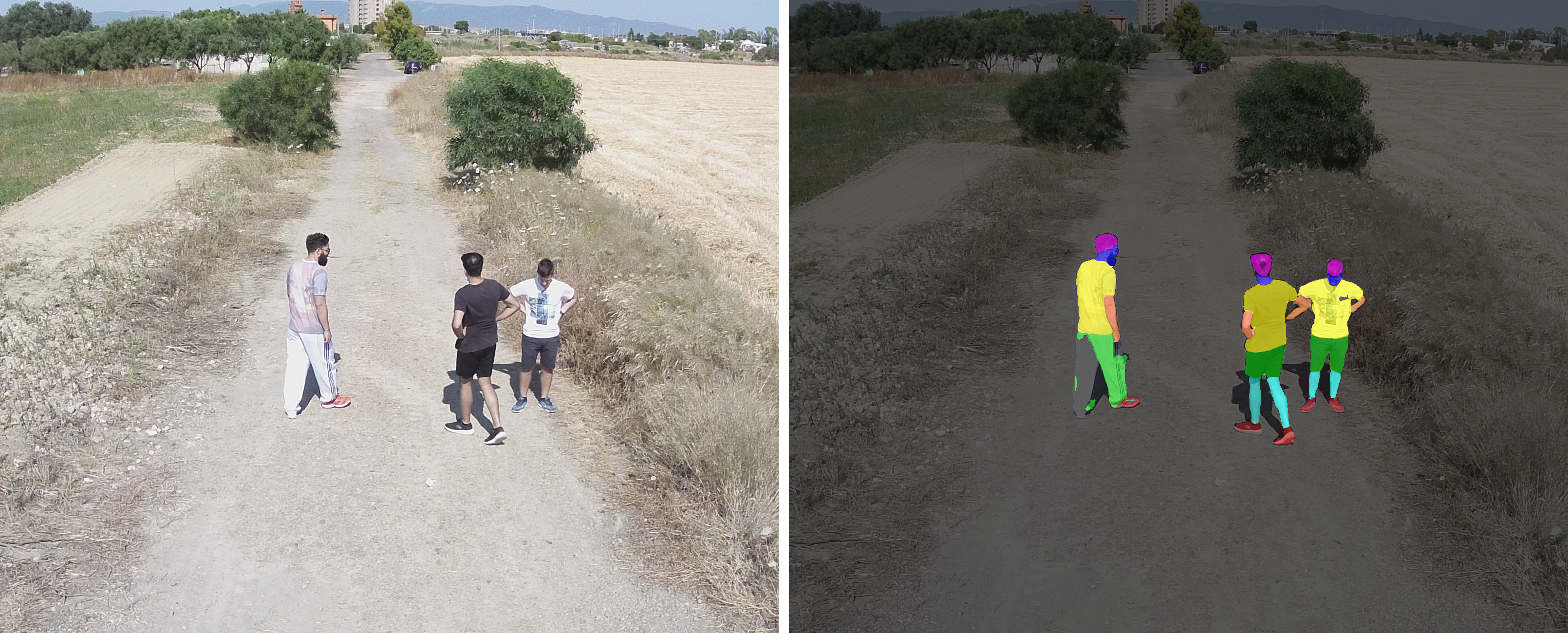}}
\caption{Example of human body segmentation model output}
\label{human-body-seg}
\end{center}
\vskip -0.2in
\end{figure}

\begin{figure}[ht]
\vskip 0.2in
\begin{center}
\centerline{\includegraphics[width=0.6\columnwidth]{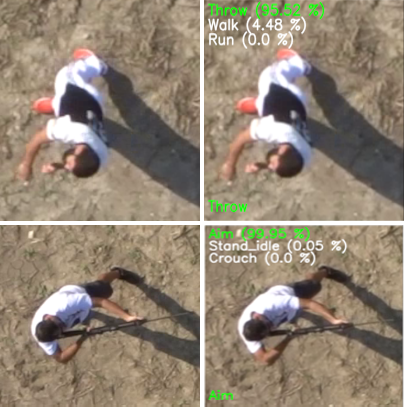}}
\caption{Example of human action recognition model output}
\label{har-output}
\end{center}
\vskip -0.2in
\end{figure}

\textbf{Figures \ref{object-detection}, \ref{road-seg}, \ref{human-body-seg} and \ref{har-output}} show four examples of the models' outputs for, respectively, object detection and localization, road segmentation, human body segmentation and human action recognition.

Video demonstrations for the four use cases are listed below:
\begin{itemize}
\item Object detection and localization: \url{https://youtu.be/XvADN61eIlw}
\item Road segmentation: \url{https://youtu.be/d-djZUXYHwg}
\item Human body segmentation: \url{https://youtu.be/lbkUOwbxiaU}
\item Human action recognition: \url{https://youtu.be/AObXkNNc-ps}
\end{itemize}

\iffalse
\begin{figure}[h]
\vskip 0.2in
\begin{center}
\centerline{\includegraphics[width=0.8\columnwidth]{images/yoloNet.png}}
\caption{YOLO base architecture schemes}
\label{object-detection-net}
\end{center}
\vskip -0.2in
\end{figure}
\fi

\iffalse
\begin{figure}[ht]
\vskip 0.2in
\begin{center}
\centerline{\includegraphics[width=\columnwidth]{images/semanticSegNet.png}}
\caption{DDRNet base architecture scheme}
\label{icml-historical}
\end{center}
\vskip -0.2in
\end{figure}
\fi

\iffalse
\begin{figure}[ht]
\vskip 0.2in
\begin{center}
\centerline{\includegraphics[width=\columnwidth]{images/harNet.png}}
\caption{Two-Stream ConvNet base architecture}
\label{icml-historical}
\end{center}
\vskip -0.2in
\end{figure}
\fi

\section{Discussion and Conclusions}
\label{conclusions}

This work presented how deep learning based computer vision algorithms have been developed, trained and deployed on an embedded NVIDIA Jetson Xavier equipped on-board a small multi-rotor drone.

All selected models have been chosen to assure a broad applicability, favoring the easiest possible integration in terms of both input sources and output third party downstream consumers. The adopted technology and the implementation choices have been driven by the goal of creating a software library able to handle different types of input data and to provide outputs that contain abstract, high level information extracted from the frames.

While here applied on RGB images coming from a visible camera, each model can work on the following type of inputs, and can easily be extended to similar ones:
\begin{itemize}
\item RGB (3-channels)/Grayscale (single-channel) images coming from visible camera
\item Gray-scale (single-channel) images coming from infra-red cameras
\item Multi-spectral and hyper-spectral images coming from ground, air or space vehicles
\end{itemize}

Each model provides its own specific output:
\begin{itemize}
\item Automatic target detection, classification and localization:
\begin{itemize}
\item One array of detections having the whole list of targets objects identified in every frame
\item One bounding box for each detection, defining target position inside the frame
\item One classification label for each detection, representing the class in which the identified and localized target belongs
\item One classification confidence for each detection, measuring the probability associated with the label assigned to the target identified in the frame
\end{itemize}
\item Context semantic segmentation (for both road segmentation and human body segmentation):
\begin{itemize}
\item One pixel map, where each frame pixel is associated with a classification label
\end{itemize}
\item Human action recognition:
\begin{itemize}
\item One classification label for each frame sequence, representing the class the target action belongs to
\item One classification confidence for each action detection, measuring the probability associated with the label assigned to the target action identified in the frame
\end{itemize}
\end{itemize}

Results presented demonstrate the successful application of state-of-the-art, deep learning based computer vision algorithms for quasi-real-time sensor processing on-board of multi-rotor drones. Performances obtained proved that this approach can be a very promising way to pursue scalability in UAVs applications, and these functionalities can be directly applied in enabling fully autonomous systems.

\clearpage

%\section*{Acknowledgments}

%To be completed for the final version.

%\section*{Resources}
%\noindent Repository: \url{https://github.com/diambra/arena}\\
%Documentation: \url{https://docs.diambra.ai}\\
%Website: \url{https://diambra.ai}\\



%%       \citet{<label>} ==> Jones et al. [21]
%%       \citep{<label>} ==> [21]
%%

\newpage
  \bibliographystyle{elsarticle-harv} 
  \bibliography{nais}


%\begin{thebibliography}{00}
%
%%% \bibitem[Author(year)]{label}
%%% Text of bibliographic item
%\end{thebibliography}
\end{document}